\documentclass[conference]{IEEEtran}
\IEEEoverridecommandlockouts
\usepackage{cite}
\usepackage{amsmath,amssymb,amsfonts}
\usepackage{algorithmic}
\usepackage{graphicx}
\usepackage{textcomp}
\usepackage{xcolor}

\usepackage[utf8]{inputenc} % allow utf-8 input
\usepackage[T1]{fontenc}    % use 8-bit T1 fonts
\usepackage{hyperref}       % hyperlinks
\usepackage{url}            % simple URL typesetting

\usepackage{booktabs}       % professional-quality tables
\usepackage{amsthm,amsmath,amsfonts}
\usepackage{mathtools}      % blackboard math symbols
\usepackage{nicefrac}       % compact symbols for 1/2, etc.
\usepackage{microtype}      % microtypography
\usepackage{graphicx}
\usepackage{subfigure}
\usepackage{fourier}
\usepackage{multirow}
\usepackage[ruled,vlined,linesnumbered]{algorithm2e}
\usepackage[english]{babel}

\def\BibTeX{{\rm B\kern-.05em{\sc i\kern-.025em b}\kern-.08em
    T\kern-.1667em\lower.7ex\hbox{E}\kern-.125emX}}
\begin{document}

\title{A Comparative study of Hyper-Parameter Optimization Tools\\
}

\author{\IEEEauthorblockN{Shashank Shekhar}
\IEEEauthorblockA{\textit{Head, AI Labs, Subex Ltd.} \\
Bangalore, India \\
shashank.shekhar@subex.com}
\and
\IEEEauthorblockN{Adesh Bansode}
\IEEEauthorblockA{\textit{Data Scientist, AI Labs, Subex Ltd.} \\
Bangalore, India \\
adesh.bansode@subex.com}
\and
\IEEEauthorblockN{Asif Salim}
\IEEEauthorblockA{\textit{Research Scientist, AI Labs, Subex Ltd.} \\
Bangalore, India \\
asif.salim@subex.com}
 
}

\maketitle

\begin{abstract}
Most of the machine learning models have associated
hyper-parameters along with their parameters. While
the algorithm gives the solution for parameters, its utility for
model performance is highly dependent on the choice of hyperparameters.
For a robust performance of a model, it is necessary
to find out the right hyper-parameter combination. Hyper-parameter
optimization (HPO) is a systematic process that
helps in finding the right values for them. The conventional
methods for this purpose are grid search and random search and
both methods create issues in industrial-scale applications.
Hence a set of strategies have been recently proposed based
on Bayesian optimization and evolutionary algorithm principles
that help in runtime issues in a production environment and
robust performance. In this paper, we compare the performance
of four python libraries, namely Optuna, Hyper-opt, Optunity
, and sequential model-based algorithm configuration (SMAC)
that has been proposed for hyper-parameter optimization. The
performance of these tools is tested using two benchmarks.
The first one is to solve a combined algorithm selection and
hyper-parameter optimization (CASH) problem The second one
is the NeurIPS black-box optimization challenge in which a multilayer
perceptron (MLP) architecture has to be chosen from a set
of related architecture constraints and hyper-parameters. The
benchmarking is done with six real-world datasets. From the
experiments, we found that Optuna has better performance for
CASH problem and HyperOpt for MLP problem.
\end{abstract}


\section{Introduction}

Machine learning algorithms for the learning tasks such
as classification, regression, clustering, etc., are associated
with parameters and hyper-parameters. The parameters
associated with the algorithm are those that can
be learned through optimization of a loss function
or through the gradients. For example, the weights and
bias associated with a linear regression can be learned by
optimizing a squared loss function. On the other hand, the
hyper-parameters are the ones that control the learning
process, and they cannot be inferred like the parameters
during the model fitting or loss function optimization. As an
example, the regularization constant in ridge regression is
a term that makes a trade-off between the empirical error
and the generalization capability of the model. It cannot
be learned via the gradients like the parameters of the
regression rather it exerts control over the entire learning
process. Hence, we can see that learning the parameters alone is not good enough for the model performance, but it
is also dependent on the right choice of hyper-parameters.
There have been studies that have shown that the set of
optimal hyper-parameters improves the performance of the
model \cite{1} \cite{2}.

Different algorithms may have different sorts of hyper-parameters
and their influence on model performance can
also, be in different ways. For example, in the case of random
forest algorithm, number of estimators, and depth of the
trees are hyper-parameters that can have a profound influence
in the model performance while minimal cost complexity
pruning parameter can have its effect depending on the
noise content in the data. In this context, the selection
of hyper-parameters in most of the algorithms is done by
putting the analyst in the loop. But this can be a costly process if we have to select the models from a collection
of algorithms as they are highly sensitive to the selection
of hyper-parameters. To carry out the tuning process in
a systematic manner considering the cost of time and
performance guarantee, we have different hyper-parameter
optimization techniques.

In this paper, we compare the  hyper-parameter optimization techniques based on Bayesian optimization (Optuna \cite{12}, HyperOpt \cite{10}) and SMAC \cite{11a}, and evolutionary or nature-inspired algorithms such as  Optunity \cite{15}. As part of the experiment, we have done a CASH \cite{3} benchmarking and the replication of NeurIPS black box optimization challenge of 2020 \cite{ID}. In the CASH problem,  12 different classifier models are chosen for solving large-scale machine learning problems, and to get the best classifier from that when applied to real-world datasets. A total of 58 hyper-parameters of 12 classifier models are tuned for the selection of the best model. For evaluation, we take accuracy and time consumption. NeurIPS black-box optimization challenge was a competition conducted to find out the best hyper-parameter optimization methods. The task was to find out an optimal architecture of a multi-layer perceptron by optimizing the hidden layer sizes, learning parameter, batch size, etc.

The contribution of this work is a comparative study of
the different tools for hyper-parameter optimization on real-world
problems in terms of the performance and runtime
characteristics.

%proves that the Bayesian optimization method using Optuna is a very efficient method for model selection and hyper-parameter optimization, and it performs better than other approaches.

\section{Hyper-parameter optimization – the formal definition} \label{sec:2}

Consider a machine learning model $\mathcal{M}$. We assume that the corresponding learning algorithm $\mathcal{A}$ associated with $\mathcal{M}$ is parameterized by a set of hyper-parameters $x = \{h_1,h_2,\dots,h_n \}$ where each $h_i \in \mathcal{X}_i$. Let $\lambda = \mathcal{X}_1 \times \mathcal{X}_2 \times \dots \times \mathcal{X}_n$  be the hyper-parameter space from which the algorithm $\mathcal{A}$ chooses its hyper-parameters. We denote this model as $ \mathcal{M} := \mathcal{A}_\lambda$.  The dataset available for the training the model is assumed to be $D = \{D_{train}, \;D_{test}\}$ where $D_{train}$ is the training data using which the model is trained and $D_{test}$ is the testing data in which the model performance is evaluated. For defining the hyper-parameter optimization problem, $\mathcal{M}_\lambda$ is assumed to be optimizing a loss function as part of its training. 

With this setting, the hyper-parameter optimization problem is to maximize the function,

\begin{equation} \label{e1}
	f(x) = l(\mathcal{A}_\lambda, D_{train}, D_{test}).
\end{equation}

Note that here $f$ is a back box function where we do not have knowledge about its analytical form.

\subsection{Challenges in Hyper-parameter tuning}
\textit{Time consumption:} If the modeling problem has the time or
scalability constraints, we shall have an efficient mechanism
for hyper-parameter tuning that process only through those
configurations that are likely to give better performance.
The conventional methods like grid search and random search may not guarantee this behavior but the Bayesian
optimization strategies have inbuilt capabilities to make
use of the posterior probability for selecting a feasible
configuration.

\textit{Variety and inter-dependency:} The hyper-parameters can
be of the continuous type that gets values in a range, categorical
type, or a discrete setting of fixing the number of layers in a
neural network. Apart from this, a choice of hyper-parameter
can be dependent on another one. They put additional
constraints on the optimization problem.

\section{Background and related works}
In this section, we review the hyper-parameter optimization techniques and other related works. 

\subsection{Grid Search}

The grid search is a technique that has been applied
classically by checking all the possible parameter
combinations. In grid search, the entire parameter space
is considered and the space is divided as in the form of
a grid. Then each of the points in the grid is evaluated as
hyper-parameters. The method can be implemented easily and simply \cite{4}. %For the experiments in this paper,  GridSearchCv implementation in scikit-learn is utilized \cite{5}. 

\textit{Limitations:} The method can be used only in the case
of a low dimensional hyper-parameter space, that is, 1-D,
2-D, etc. The method is time-consuming for a larger number of
parameters. The method cannot be applied for model selection as it can be used only for tuning a single model.

\subsection{Random Search }
Random search is another commonly used approach in which the hyper-parameters are selected at random, independent of other choices. The method is simple to implement and it is well suited for learning gradient-free function. Compared to the grid search, the random search method converges faster. The method finds an optimal model by effectively searching a larger, less promising hyper-parameter space \cite{4}.
%For the experiments in this paper,   RandomizedSearchCV implementation in scikit-learn is utilized \cite{5}.  

\textit{Limitations:} It is a time-consuming method as the evaluation of the function gets expensive. The method does not have the capability of model selection as it is used only with a single model. Compared to Bayesian optimization, this method does not exploit the knowledge of well-performing search space \cite{5} \cite{6}. 

\subsection{Bayesian Hyper-parameter Optimization}
In this section, we describe briefly the techniques involved in Bayesian hyper-parameter optimization.

\subsubsection{Overview}
The hyper-parameter optimization problem given in \eqref{e1} involves a black-box function $f$ whose analytical form is not known. We do not have the knowledge about its derivatives with respect to the hyper-parameters and it may not be convex. Due to these reasons, it is very difficult to apply the classical optimization problems. It is in this context we go for Bayesian optimization \cite{7}.

We assume that we have certain samples from the function $f$, that is, we know for certain hyper-parameter combination what is their performance as a prior information. We denote $D_{t} = \Big\{\big(x_1,f(x_1)\big), \big(x_2,f(x_2)\big) \dots, \big(x_t,f(x_t)\big) \Big\}$ as $t$ number of such samples. Note that here each $x_i$ corresponds to a particular hyper-parameter configuration. It is also assumed that $f$ follows a prior distribution which is $P(f)$. Hence the posterior distribution can be written as,

\begin{equation} \label{e2}
	P(f|D_{t}) \propto  P(D_{t}|f) P(f).
\end{equation}

\begin{figure}[h]
	\centering
	\includegraphics[scale=0.75]{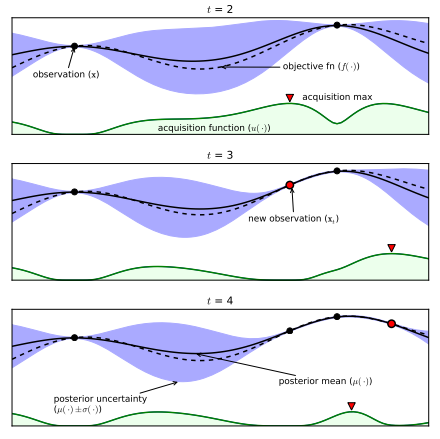}
	%\fbox{\rule[-.5cm]{0cm}{4cm} \rule[-.5cm]{4cm}{0cm}}
	\caption{A demonstration of Bayesian optimization on a 1-dimensional maximization problem.
		The figures show a Gaussian process (GP) approximation of the objective function over
		four iterations of sampled values of the objective function. The figure also shows the
		acquisition function in the lower shaded plots. The acquisition is high where the GP
		predicts a high objective (exploitation) and where the prediction uncertainty is high
		(exploration) -- areas with both attributes are sampled first. Note that the area on the
		far left remains unsampled, as it has high uncertainty and it is (correctly) predicted
		to offer little improvement over the highest observation \cite{8}]. }
	\label{f1}
\end{figure}

This posterior distribution helps to make a better estimate of the hyper-parameter configuration given the performance of observed ones \cite{9}. In practice, the objective function $f$ is evaluated with a \textit{surrogate function}. The next point at $t$+1 to be evaluated is found out using an \textit{acquisition function}. The process is graphically visualized in Figure \ref{f1}. The \textit{acquisition function} adheres to the property of exploring in the regions where the objective function is uncertain and exploiting the values of $\lambda$  where the objective function has the minimum. Hence it is likely to provide a new candidate configuration that is better than the previous ones by finding a balance between exploration and exploitation.

The Bayesian optimization technique takes only a few
evaluations in the optimization problem.  This is particularly useful in the case where the cost of the function evaluation is very high. In the following sections, we briefly discuss the \textit{surrogate} and \textit{acquisition functions} which are the key components in Bayesian optimization.

\subsubsection{Surrogate functions}

The common choices of the surrogate function are Gaussian processes and Tree-structured Parzen Estimators (TPE).

\textbf{Gaussian Process (GP):} A GP is a distribution over functions parameterized by a mean function $m$ and a covariance function $k$. That is, $f(x) \sim GP\big(m(x), k(x,x')\big).$ 
A common choice for the covariance function is the squared exponential function, $k(x_i, x_j) = \text{exp} \Big( -\frac{1}{\beta} \Vert x_i - x_j \Vert^2\Big)$ where $\beta$ is a hyper-parameter.

Considering the prior observations from $D_t$, the function values $f$ are drawn from a normal distribution $\mathcal{N}(0,K)$, where $K$ is the covariance matrix.
%\[
%K = \begin{bmatrix} 
%	k(x_1,x_1) & k(x_1,x_2) & \dots & k(x_1,x_t) \\
%	\vdots & \vdots & \vdots & \vdots\\
%	k(x_t,x_1) &  k(x_t,x_2)  & \dots    & k(x_t,x_t) 
%\end{bmatrix}
%\]
Let the new point of evaluation returned by the \textit{acquisition function} be $x_{t+1}$. We shall now estimate $f_{t+1}$ using the posterior described by \eqref{e2}. In that case, by the properties of GP,

\begin{gather*}
	\begin{bmatrix} f_{1:t}  \\   f_{t+1} \end{bmatrix}
	\sim \mathcal{N} \Biggl( 
	0, 
	\begin{bmatrix}
		K &
		\textbf{k} \\
		\textbf{k}^T &
		k(x_{t+1},x_{t+1})
	\end{bmatrix}
	\Biggr)
\end{gather*}

where $f_{1:t} = \big[f(x_1),f(x_2),\dots,f(x_t)\big]$, and $\textbf{k} = \big[k(x_{t+1},x_1), k(x_{t+1},x_2), \dots k(x_{t+1},x_t)\big]$.

Now the posterior distribution can be written as:
\begin{equation*}
	P(f_{t+1}|D_t,x_{t+1}) = \mathcal{N}(\mu_{t+1}, \sigma_{t+1}^2)
\end{equation*} 

where $\mu_{t+1} = \textbf{k}^T K^{-1} f_{1:t}$ and
$\sigma_{t+1}^2 = k(x_{t+1}, x_{t+1}) - \textbf{k}^TK^{-1}\textbf{k}$. 

The choices for covariance function are Automatic relevance
determination (ARD) kernels and Matern kernels. The alternatives for GP are random forest and tree parzen estimators. 

\subsubsection{Acquisition functions}
The purpose of the acquisition function is to choose the next point of evaluation of the function $f$ such that the optimization shall progress towards the maxima. This is accomplished by finding the argument point that maximizes the acquisition function.

One of the popular choices is to maximize the \textit{expected improvement (EI)} with respect to $f(x^+)$ where $x^+ = \text{arg max}_{x_i \in x_{1:t}} f(x_i)$. The \textit{improvement} function is defined as, $I(x) = \text{max}\big\{0, f_{t+1}(x) - f(x^+)\big\}$. Now the new point for evaluation is found out by maximizing the EI, $x = \underset{x}{\text{arg max}}\; \mathbb{E}\big( I(x) | D_t \big).$ 

The likelihood of $I$ is computed from the normal density function as, $
	\frac{1}{\sqrt{2\pi} \sigma(x)} \text{exp} \left( - \frac{\big(\mu(x) - f(x^+) - I\big)^2}{2\sigma^2(x)}\right)$.

With this setting, the expected improvement can be evaluated as follows,
\begin{equation} \label{kdelta}
	EI(x) =\left\{
	\begin{array}{ll} 
		\big(\mu(x) - f(x^+)\big) \Phi(Z) + \sigma(x)\phi(Z) \; \text{if } \sigma(x) > 0
		\\ 0, \; \text{if } \sigma(x) = 0  \\
	\end{array}
	\right.
\end{equation}
where $Z = \big(\mu(x) - f(x^+)\big)/\sigma(x)$, $\phi$ and $\Phi$ denote the probability distribution function and cumulative distribution function of the standard normal distribution respectively. 

The discussion about the evolutionary algorithms is omitted for brevity. 
\section{Hyper-parameter Optimization Tools }
This section briefly describes the tools we have used.

\subsection{HyperOpt}
HyperOpt \cite{10} \cite{11} is a python library that works based on the sequential model-based optimization (SMBO) \cite{11a} associated with the Bayesian optimization defined in the previous section. It provides an interface where the user can configure the search space of the variables, customizing an evaluation function and assigning the loss function for evaluating the points in search space. It can also facilitate to set up a single large hyperparameter optimization problem in which a variety of algorithms, data pre-processing modules and their hyper-parameters are combined.

The main components of HyperOpt are a search
domain, an objective function, and an optimization algorithm.
The search domain provides greater flexibility in
the optimization process as it can be characterized by
continuous, ordinal, or categorical variables. The objective
function can be a user-defined python function that accepts
the variables and return a loss function corresponding to
that variable combination. The search algorithm stands for
the surrogate function defined in the previous section. The
choices provided by the tool are random search, TPE, and
adaptive TPE. The tool also provides the facility for parallel
implementation utilizing Apache Spark and MongoDB.

\subsection{Optuna}
The features of Optuna as quoted by the authors are \textit{" (1) define-by-run API that allows users to construct the parameter search space
	dynamically, (2) efficient implementation of both searching and pruning strategies, and (3) easy-to-setup, versatile
	architecture that can be deployed for various purposes, ranging from scalable distributed computing to light-weight
	experiment conducted via interactive interface."}\cite{12}

The define-by-run API is unique to Optuna whereas, in the case of HyperOpt, the users have to specify the search spaces for each hyper-parameters explicitly. The objective function of Optuna receives \textit{a trial object} that is embedded with the parameter space and function to optimize instead of the hyper-parameter values. The define-by-run API also gives the ability for modular programming.%, as an example, the topology of a neural network and the parameters associated with gradient descent learning algorithm can be incorporated as two separate sub-routines into  the Optuna objective function.

In the hyper-parameter optimization problem, it is required to deal with relational sampling as well as independent sampling. Relational sampling refers to the correlation that exists between the parameters while the other refers to the sampling process that performs independently. Optuna has the ability to identify the experiment runs that contain these concurrence relations. %It also provides the facility to use the relations along with the user specified relational sampling algorithms such as the covariance matrix adaption evolutionary strategy (CMA-ES).   

Optuna provides the pruning feature that helps to prematurely terminate the runs that are not optimal. For this purpose, the intermediate objective values are monitored and those that do not meet predefined conditions are terminated. This is enabled by an asynchronous successive halving algorithm. It also provides the distributed computing capabilities.

\subsection{Optunity} 
Optuniy is a framework for CASH problem \cite{3} with a set of different solvers. Optunity provides a collection of solvers -  basic solver, grid search, random search, evolutionary methods such as particle swarm optimization (PSO) and covariance matrix
adaption evolutionary strategy (CMA-ES). In the experiments related to this paper, only  PSO is used. 
The features of Optunity are it can be configured with minimal efforts, number of evaluations can be set to an upper limit and the hyper-parameters can be given box constraints \cite{15}. 

%In Optunity,  categorical hyper-parameters cannot be given directly in search space. For categorical hyper-parameter, we must transform to integer hyper-parameters (by indexing), and these integers (index) hyper-parameter are treated as continuous hyper-parameters \cite{15}. %In our experiment, we drop the categorical hyper-parameters to remove complexity. This is a drawback of Optunity compared to another approach.

\subsection{Sequential model based algorithm configuration (SMAC)}

SMAC is \cite{11a} introduced as an improvement to the classical  sequential model-based optimization (SMBO). SMAC has an efficient sampling mechanism in which multiple instances can be processed by considering their performance  into account. It has also introduced the mechanism to incorporate categorical parameters into the optimization procedures.

A model to predict the runtime of the algorithm for different parameter configurations is embedded in SMAC. It helps to reduce the number of runs of the algorithms. SMAC also has the capabilities to select promising configurations in
large mixed numerical/categorical configuration spaces.
\begin{table}
	\caption{Dataset details }
	\label{t2}
	\centering
	\scalebox{1}{
		\begin{tabular}{llllll}
			
			\toprule
			%\multicolumn{2}{c}{Part}                   \\
			%\cmidrule(r){1-2}
			Dataset    & Classes & Size &  Num.feat. & Cat.feat.\\ %& Description\\
			\midrule

			dna & 3 & 3186 &	0 & 181 \\ %& all features categorical/high dimensional/small size\\
			
			Electricity & 2 &	45312 &	7 &	2 \\ %& low dimensional/large size\\
			
			Gas drift & 6 &  13910 &  128 	& 1 \\ %& 	  multi class/moderately high dimensional/large size \\
			
			Nomao & 2 &	34465 &	89 & 30 \\ %&  have numerical \& categorical features/ large size\\
			
			Pendigits &  10 & 10992 &	16 & 1 \\ %& multi class/low dimensional\\
			
			Semion & 10 & 1593 & 256 & 1  \\ %& high dimensional/low size\\
			
			\bottomrule
	\end{tabular}}
\end{table}

\section{Experiments}
In this section, we describe the details of the experiments, datasets used, system setup, implementation details and metrics used for the classification tasks. 

\subsection{Datasets}
We have taken six publicly available datasets from OpenML \cite{13} machine learning repository. The details of the datasets are described in Table \ref{t2}. The datasets are chosen in a way that they cover the probable real-world scenarios that commonly occur. %The description of the dataset is given to understand these scenarios.

\subsection{System setup and Implementation}

We built a model of classifiers in the Python3 framework. The system configuration is 3.50 GHz  Intel Xeon i7-4771 CPU with 32GB RAM. %We used  Scikit-learn with different optimization tools such as GridSearchCV, RandomSearchCV along with HyperOpt, Optuna, and Optunity. 
We use K-fold cross-validation (K = 3) and compute the f1-score as the performance metric. The methods are evaluated for 50 iterations.
%The evaluation score used in the experiment is f1-score.

%\subsection{Evaluation metrics}
%In this paper, we use the f1-score as an evaluation metric of our model. The f1-score combines precision and recall. It can be interpreted as a weighted average of the recall and precision. Therefore, it takes both false negative and false positives into account. The f1-score is more useful than accuracy in classification, especially if we have uneven class distribution. If the weight of false negative and false positive are very different, then both precision and recall work better. Accuracy works better if the weight of false negative and false positive are similar. The f1-score reaches its best value at 1 and the worst score at 0 and it is defined as the harmonic mean of precision and recall, that is,

%\begin{equation*}
%	\text{f1-score} = 2*\frac{\text{precision * recall}}{\text{precision + recall}}
%\end{equation*}

%where precision = (true positive)/(true positive + false positive) and recall = (true positive)/(true positive + false negative). The f1-score and the runtime for each of the methods are recorded for the comparison purposes.

\subsection{Benchmark-I}

The configuration space for the CASH algorithm contains 12 classifiers with a total number of 58 hyper-parameters. This configuration space is listed in Table \ref{t1}. %If algorithm not supporting condition configuration spaces use it without conditional dependencies. 

\begin{table}
	\caption{Configuration space for classification algorithms. The number of categorical, continuous, and total number of hyperparameters is listed. }
	\label{t1}
	\centering
	\scalebox{1}{
		\begin{tabular}{llllll}
			
			\toprule
			%\multicolumn{2}{c}{Part}                   \\
			%\cmidrule(r){1-2}
			Sl.No    & Algorithm & Cat. & Cont. & Total\\
			\midrule
			1 &  Bernoulli naive Bayes & 1 &	1 & 2 \\
			
			2 & Multinomial naive Bayes & 1 &	1 & 2 \\
			
			3 & Decision Tree &	1 &	3 &	4 \\
			
			4 & Extra Trees	& 2 & 	3	& 5 \\
			
			5 & Gradient Boosting&	1 &	5 &	6  \\
			
			6 & Random Forest &	2 &	4 &	6   \\
			
			7 &  K Nearest Neighbors & 2 &	1	& 3 \\
			
			8 & Logistic Regression &	3 &	1	& 4   \\
			
			9 & Linear SVM &	2 &	2 &	4 \\
			
			10 & SGD Classifier	& 4 & 	6	& 10 \\
			
			11 & XGB Classifier &	1 &	5 &	6  \\
			
			12 & LGBM Classifier &	0 &	6 & 6   \\
			\bottomrule
	\end{tabular}}
\end{table}

\subsubsection{Results and Observations}

The results are given in Table \ref{t3}. We can see that in terms of time, \textit{Optuna} has the better performance in all the datasets except \textit{semeion}. However, this pattern is not followed in terms of the performance score. In the case of \textit{dna, nomao, and pendigits}, the score of all the methods is satisfactory. But in the case of \textit{electricity} and \textit{gas drift}, \textit{Optunity PSO} has a high score although the time taken is higher than \textit{Optuna}. Note that these datasets have a high sample size and moderately high dimensional. On the other hand, the score of \textit{Optunity PSO} is lower in the high dimensional \textit{semeion} dataset although it has the least run time. In this case, \textit{Optuna} has a high score with a slight increase in runtime compared with \textit{Optunity PSO}. 

Comparing \textit{HyperOpt} and \textit{SMAC}, their performance score is similar in all datasets. But the runtime of \textit{SMAC} is higher in most of the datasets. The performance score of \textit{HyperOpt} and \textit{Optuna} is similar in all datasets. Comparing the runtime, \textit{Optuna} is better in most of the datasets except \textit{semeion} although the difference is not significant.

Another useful comparison is about the \textit{random} search and searches with \textit{TPE} of \textit{HyperOpt} and \textit{Optuna}. We can see that the performance of \textit{TPE} is slightly better compared to \textit{random} search.

From the experiments, we can observe that considering the trade-off between the performance score and runtime, which is a common scenario in industry applications, \textit{Optuna} is a better choice for HPO  compared to other approaches. In the case of very large data, \textit{Optunity} can also be a good choice and its runtime can be further optimized through parallel computation.
\begin{table}[]
	\centering
	\caption{Result of Benchmark-I}
	\label{t3}
	\begin{tabular}{@{}llrr@{}}
		\toprule
		\textbf{Dataset}             & \textbf{Method} & \multicolumn{1}{l}{\textbf{Score}} & \multicolumn{1}{l}{\textbf{Time}} \\ \midrule
		\multirow{6}{*}{dna}         & HyperOpt Random & 0.9538178                       & 366   \\
		& HyperOpt TPE    & 0.9579369                       &  322   \\
		& Optuna Random   & 0.9580583                       &  94   \\
		& Optuna TPE      & 0.9602619                       &  521    \\
		& Optunity PSO    & 0.925369                        &  200    \\
		& SMAC            & 0.9595218                        &  807    \\
		\hline
		\multirow{6}{*}{electricity} & HyperOpt Random & 0.6969229                      &  546      \\
		& HyperOpt TPE    & 0.7011213                        &  341       \\
		& Optuna Random   & 0.6942903                       &  494     \\
		& Optuna TPE      & 0.7011213                       &  167     \\
		& Optunity PSO    & 0.9074971                      &  190   \\
		& SMAC            & 0.7011213                       &  393    \\
		\hline
		\multirow{6}{*}{gas drift}   & HyperOpt Random & 0.850121                           & 23869                          \\
		& HyperOpt TPE    & 0.88168                            & 14105                         \\
		& Optuna Random   & 0.868639                           & 14549                        \\
		& Optuna TPE      & 0.882114                           & 4403                        \\
		& Optunity PSO    & 0.982105                           & 11923                         \\
		& SMAC            & 0.825001                           & 12966                        \\
		\hline
		\multirow{6}{*}{nomao}       & HyperOpt Random & 0.926516                           & 1711                        \\
		& HyperOpt TPE    & 0.931569                           & 1441                       \\
		& Optuna Random   & 0.93677                            & 1064                          \\
		& Optuna TPE      & 0.937144                           & 1124                         \\
		& Optunity PSO    & 0.938943                           & 1554                          \\
		& SMAC            & 0.937003                           & 2622                         \\
		\hline
		\multirow{6}{*}{pendigits}   & HyperOpt Random & 0.979188                           & 1783                        \\
		& HyperOpt TPE    & 0.987569                           & 1012                        \\
		& Optuna Random   & 0.982508                           & 336                         \\
		& Optuna TPE      & 0.983886                           & 897                          \\
		& Optunity PSO    & 0.958028                           & 2980                         \\
		& SMAC            & 0.984226                           & 860                            \\
		\hline
		\multirow{6}{*}{semeion}     & HyperOpt Random & 0.901575                           & 762                          \\
		& HyperOpt TPE    & 0.915246                           & 458                          \\
		& Optuna Random   & 0.92832                            & 895                           \\
		& Optuna TPE      & 0.934616                           & 1989                         \\
		& Optunity PSO    & 0.700295                           & 538                         \\
		& SMAC            & 0.929794                           & 1647                         \\ 
		\bottomrule
	\end{tabular}
\end{table}

\subsection{Benchmark-II}
As part of the NeurIPS black-box optimization challenge, the task was to choose a multi-layer perceptron (MLP) architecture by optimizing through a set of hyper-parameters as described in Table \ref{t4}. 

\begin{table}[]
		\caption{Configuration space for MLP tuning. The type, nature of search space and the range of values for searching are given as columns.}
	\label{t4}
	\centering
	\begin{tabular}{llll}
		\toprule
		\textbf{Hyper-parameter} & \textbf{Type} & \textbf{Space} & \textbf{Range}    \\
		\midrule
		hidden layer size        & integer       & linear         & (50, 200)         \\
		alpha                    & real          & log            & (10$^{-5}$, 10)       \\
		batch size               & integer       & linear         & (10, 250)         \\
		learning rate            & real          & log            & (10$^{-5}$, 10$^{-1}$)     \\
		tolerance                & real          & log            & (10$^{-5}$, 10$^{-1}$)      \\
		validation fraction      & real          & real           & (0.1, 0.9)        \\
		beta 1                   & real          & logit          & (0.5, 0.99)       \\
		beta 2                   & real          & logit          & (0.9, 1.0-10$^{-6}$) \\
		epsilon                  & real          & log            & (10$^{-9}$, 10$^{-6}$)     \\
		\bottomrule
	\end{tabular}
\end{table}

% \usepackage{multirow}
\begin{table}[]
		\centering
	\caption{Result of Benchmark-II}
	\label{t5}
	\begin{tabular}{llll}
		\toprule
		\textbf{Dataset}             & \textbf{Method} & \textbf{Score} & \textbf{Time} \\
		\midrule
		\multirow{6}{*}{dna}         & HyperOpt Random & 0.942171       & 72       \\
		& HyperOpt TPE    & 0.946809       & 87        \\
		& Optuna Random   & 0.947759       & 267     \\
		& Optuna TPE      & 0.947453       & 415    \\
		& Optunity PSO    & 0.901858       & 243     \\
		& SMAC            & 0.950322       & 388      \\
		\hline
		\multirow{6}{*}{electricity} & HyperOpt Random & 0.669953       & 366      \\
		& HyperOpt TPE    & 0.683778       & 229      \\
		& Optuna Random   & 0.692635       & 2042     \\
		& Optuna TPE      & 0.696458       & 1269      \\
		& Optunity PSO    & 0.583639       & 1550     \\
		& SMAC            & 0.685409       & 2844     \\
		\hline
		\multirow{6}{*}{gas drift}   & HyperOpt Random & 0.742839       & 257   \\
		& HyperOpt TPE    & 0.847971       & 173      \\
		& Optuna Random   & 0.89409        & 2798     \\
		& Optuna TPE      & 0.908367       & 8509     \\
		& Optunity PSO    & 0.851657       & 706     \\
		& SMAC            & 0.898836       & 3732     \\
		\hline
		\multirow{6}{*}{nomao}       & HyperOpt Random & 0.909167       & 624      \\
		& HyperOpt TPE    & 0.920299       & 501      \\
		& Optuna Random   & 0.931145       & 3842     \\
		& Optuna TPE      & 0.931401       & 4933      \\
		& Optunity PSO    & 0.885599       & 1732      \\
		& SMAC            & 0.930689       & 12316      \\
		\hline
		\multirow{6}{*}{pendigits}   & HyperOpt Random & 0.967597       & 205     \\
		& HyperOpt TPE    & 0.981328       & 182      \\
		& Optuna Random   & 0.985192       & 538      \\
		& Optuna TPE      & 0.987346       & 791      \\
		& Optunity PSO    & 0.925501       & 544     \\
		& SMAC            & 0.988169       & 1060      \\
		\hline
		\multirow{6}{*}{semeion}            & HyperOpt Random & 0.918278       & 65      \\
		& HyperOpt TPE    & 0.922646       & 57       \\
		& Optuna Random   & 0.927138       & 254      \\
		& Optuna TPE      & 0.930875       & 132     \\
		& Optunity PSO    & 0.72276        & 196     \\
		& SMAC            & 0.934178       & 140    \\
		\bottomrule
	\end{tabular}
\end{table}

\subsubsection{Results and Observations}

The performance score of \textit{Optuna} is the highest for all datasets while in terms of runtime it is the \textit{HyperOpt}.  It can be noted that there is no considerable difference in the score between both the methods except in the case of \textit{gas drift} and \textit{nomao}. For a few datasets, \textit{SMAC} also has a similar performance to \textit{Optuna} but it has a comparatively higher runtime compared to other methods. 

The performance score of \textit{Optunity PSO} is the least among all the methods and its runtime performance is in between \textit{Optuna} and \textit{HyperOpt} and better than \textit{SMAC}. From the experiments, we can observe that for neural network applications \textit{HyperOpt} is the better option.

 \begin{figure}[h]
	\centering
	\includegraphics[scale=0.45]{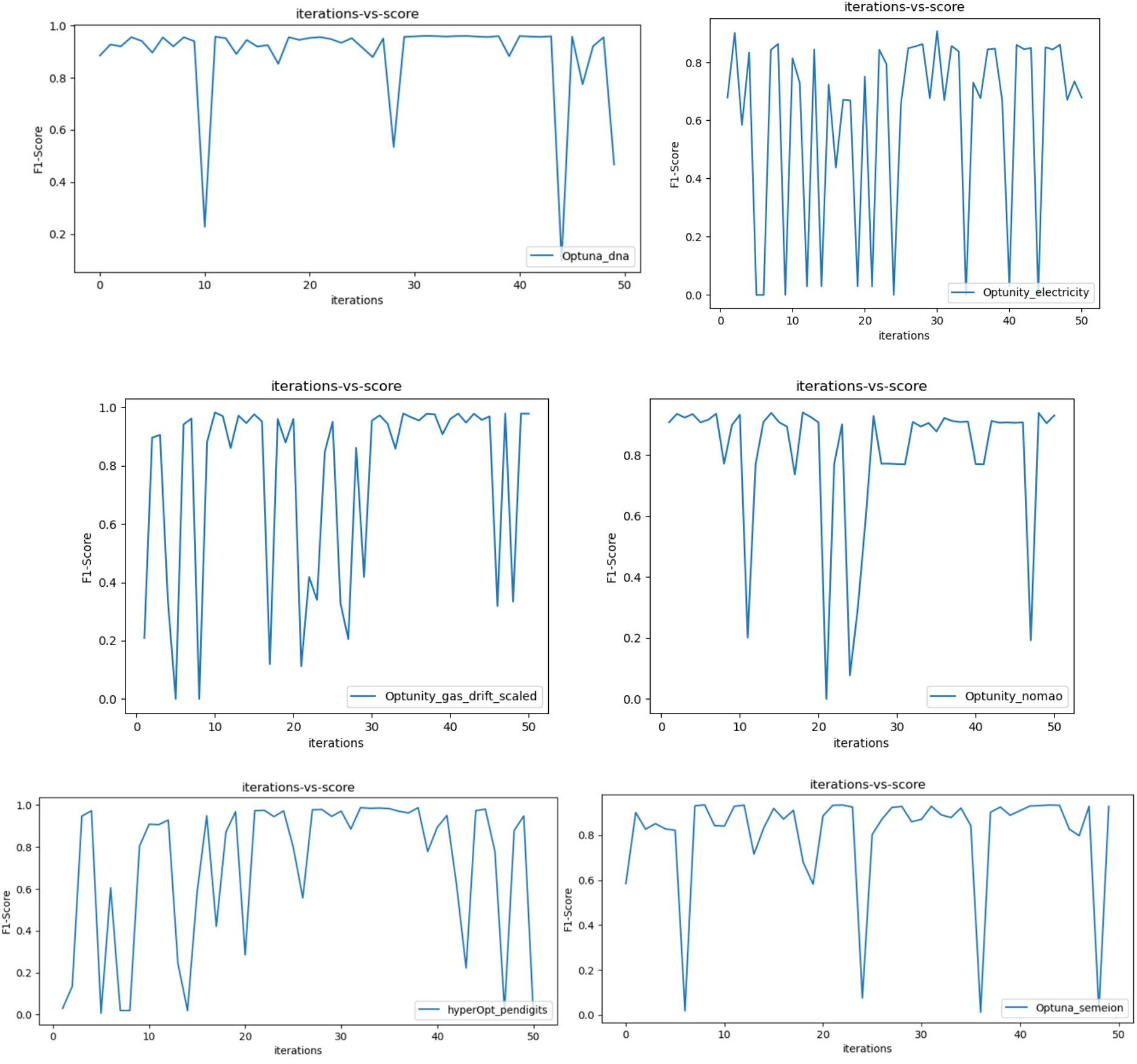}
	%\fbox{\rule[-.5cm]{0cm}{4cm} \rule[-.5cm]{4cm}{0cm}}
	\caption{Plot of  f1-score v/s iteration for benchmark-I. The corresponding algorithm and dataset are marked as a legend in each plot.}
	\label{f2}
\end{figure}
  
 \begin{figure}[h]
	\centering
	\includegraphics[scale=0.45]{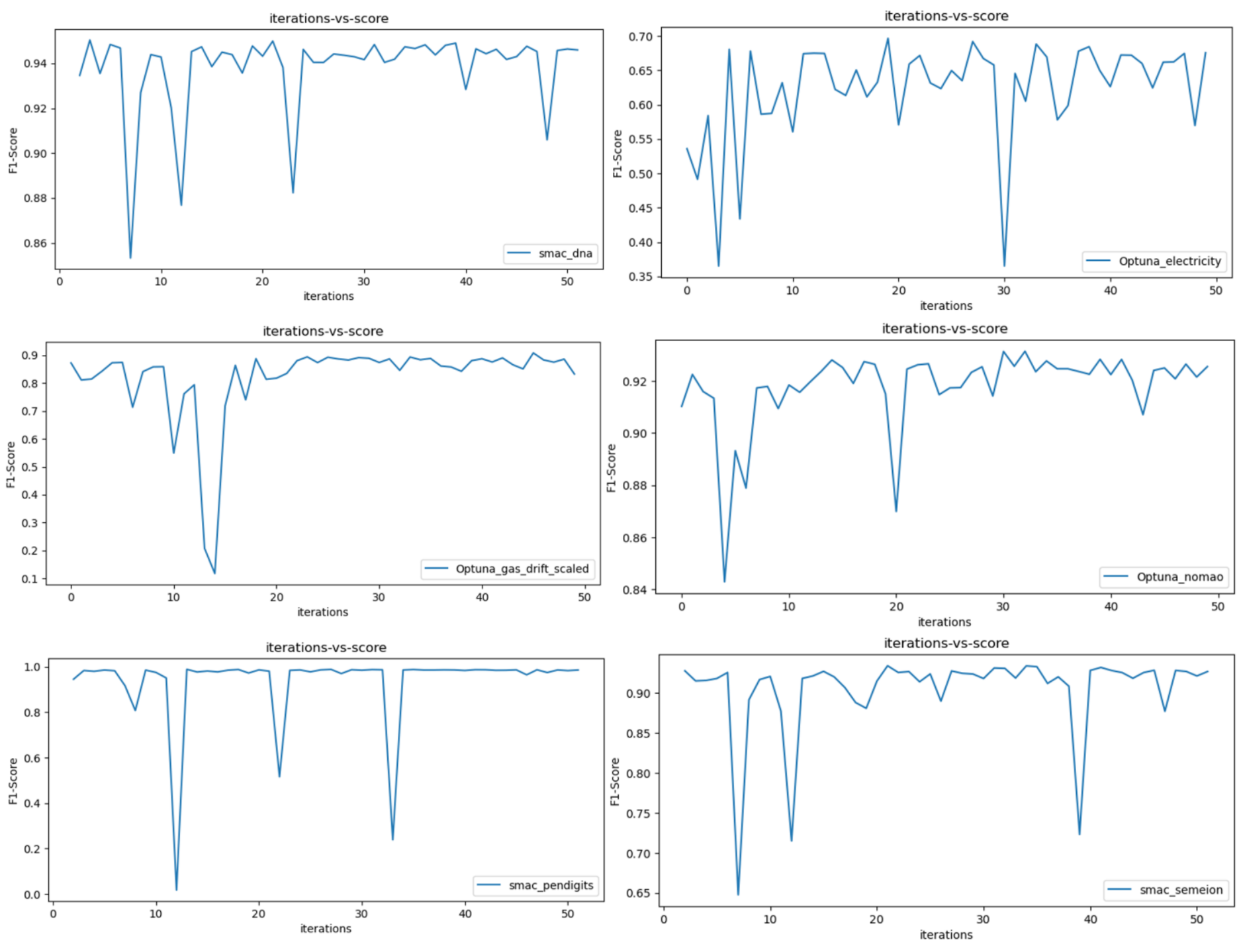}
	%\fbox{\rule[-.5cm]{0cm}{4cm} \rule[-.5cm]{4cm}{0cm}}
	\caption{Plot of  f1-score v/s iteration for benchmark-II. The corresponding algorithm and dataset are marked as a legend in each plot.}
	\label{f3}
\end{figure}
The plot of  f1-score against iteration for the best performing model for each dataset are given in Fig.\ref{f2} and Fig.\ref{f3} for the benchmarks-I and II respectively. From the figures, we can see that the Optunity and HyperOpt shows an oscillatory behavior while Optuna is more stable.
\section{Conclusion}
We compared the hyper-parameter optimization libraries HyperOpt, Optuna, Optunity, and SMAC on two benchmarks. Based on our experiments we found that for the CASH problem benchmark Optuna is the good option considering the trade-off between the runtime and performance score. On the other hand, for the MLP problem, it is the HyperOpt. 

\bibliographystyle{ieeetr}
\bibliography{reference}

\end{document}